\documentclass[lettersize,journal]{IEEEtran}
\usepackage{amsmath,amsfonts}
\usepackage{algorithmic}
\usepackage{algorithm}
\usepackage{array}
\usepackage[caption=false,font=normalsize,labelfont=sf,textfont=sf]{subfig}
\usepackage{textcomp}
\usepackage{stfloats}
\usepackage{url}
\usepackage{verbatim}
\usepackage{graphicx}
\usepackage{cite}

\usepackage{textcomp}
\usepackage{stfloats}
\usepackage{url}
\usepackage{verbatim}
\usepackage{graphicx}
\usepackage{bbding}
\usepackage{textcomp}
\usepackage{booktabs} 
\usepackage{multirow}
\usepackage{threeparttable} 
\usepackage{soul}
\usepackage{color}
\usepackage{cite}
\usepackage{bbm}
\usepackage{hyperref}

\begin{document}

\title{Uncertainty-aware Sampling for Long-tailed Semi-supervised Learning}

\author{Kuo Yang, ~\IEEEmembership{Student Member,~IEEE,}  Duo Li$^{*}$, ~\IEEEmembership{Member,~IEEE,}
Menghan Hu$^{*}$, ~\IEEEmembership{Senior Member,~IEEE,} 
Guangtao Zhai, ~\IEEEmembership{Senior Member,~IEEE,}
Xiaokang Yang, ~\IEEEmembership{Fellow,~IEEE,}
Xiao-Ping Zhang, ~\IEEEmembership{Fellow,~IEEE}
        % <-this % stops a space
\thanks{This work is sponsored by the National Natural Science Foundation of China (No. 62371189).}
\thanks{Kuo Yang, and Menghan Hu are with the Shanghai Key Laboratory
of Multidimensional Information Processing, School of Communication
and Electronic Engineering, East China Normal University, Shanghai
200241, China.}% <-this % stops a space
\thanks{Duo Li is with the Kargobot of DiDi, Shanghai 201210, China.}% <-this % stops a space
\thanks{Guangtao Zhai, and Xiaokang Yang are with the Institute of Image Communication and Network
Engineering, Shanghai Jiao Tong University Shanghai, 200240, China.}
\thanks{Xiao-Ping Zhang is with Tsinghua Berkeley Shenzhen Institute, Shenzhen,
China and the Department of Electrical, Computer and Biomedical Engineering,
Toronto Metropolitan University, ON M5B 2K3, Canada.
}
\thanks{$^{*}$Corresponding authors: Duo Li; Menghan Hu}% <-this % stops a space
}

% \markboth{IEEE TRANSACTIONS ON PATTERN ANALYSIS AND MACHINE INTELLIGENCE,~Vol.~14, No.~8, August~2023}%
% %\markboth{Journal of \LaTeX\ Class Files,~Vol.~14, No.~8, August~2023}%



\maketitle

\begin{abstract}
For semi-supervised learning with imbalance classes, the long-tailed distribution of data will increase the model prediction bias toward dominant classes, undermining performance on less frequent classes. Existing methods also face challenges in ensuring the selection of sufficiently reliable pseudo-labels for model training and there is a lack of mechanisms to adjust the selection of more reliable pseudo-labels based on different training stages. To mitigate this issue, we introduce uncertainty into the modeling process for pseudo-label sampling, taking into account that the model performance on the tailed classes varies over different training stages. For example, at the early stage of model training, the limited predictive accuracy of model results in a higher rate of uncertain pseudo-labels. To counter this, we propose an Uncertainty-Aware Dynamic Threshold Selection (UDTS) approach. This approach allows the model to perceive the uncertainty of pseudo-labels at different training stages, thereby adaptively adjusting the selection thresholds for different classes. Compared to other methods such as the baseline method FixMatch, UDTS achieves an increase in accuracy of at least approximately 5.26$\%$, 1.75$\%$, 9.96$\%$, and 1.28$\%$ on the natural scene image datasets CIFAR10-LT, CIFAR100-LT, STL-10-LT, and the medical image dataset TissueMNIST, respectively. The source code of UDTS is publicly available at: \href{https://github.com/yangk802/UDTS}{https://github.com/yangk/UDTS}.

\end{abstract}


\section{Introduction}
\IEEEPARstart{I}{n} recent years, deep neural networks\cite{lecun2015deep} have achieved remarkable success in various tasks, such as object classification \cite{gehler2009feature}\cite{he2016deep}, face recognition \cite{zhao2003face} and gesture recognition \cite{mitra2007gesture}. These achievements are largely attributed to the availability of large and balanced public datasets \cite{gupta2019lvis}\cite{van2018inaturalist}. However, the real-world data distributions often exhibit a long-tailed nature \cite{he2009learning}, where a majority of data belongs to a few head classes while tail classes contain relatively sparse data. When dealing with datasets exhibiting a long-tailed distribution, model predictions often display a bias towards dominant classes, resulting in diminished recognition of tail data and consequently lower overall accuracy. Addressing these challenges is crucial for advancing object recognition and facilitating the broader adoption of deep learning in real-world scenarios \cite{zhang2021deep}.
%\IEEEPARstart{I}{n} recent years, deep neural networks\cite{lecun2015deep} have achieved remarkable success in various tasks, such as object classification \cite{gehler2009feature}\cite{he2016deep}, face recognition \cite{zhao2003face} and gesture recognition \cite{mitra2007gesture}. These achievements are largely attributed to the availability of large and balanced public datasets \cite{gupta2019lvis}\cite{van2018inaturalist}. However, , the real-world data distributions often exhibit a long-tailed nature \cite{he2009learning}, where a majority of data belongs to a few head classes while tail classes contain relatively sparse data. In the face of datasets with long-tailed distribution, the prediction of the model tends to be biased toward most classes, the recognition effect of the tail data is relatively poor and the accuracy is also relatively low. Therefore, in the face of long-tailed distribution data\cite{zhang2021deep} in the real world, it brings great challenges to object recognition and greatly hinders the wide application of deep learning.
\begin{figure*}[tp]
    \centering
    \includegraphics[width=15cm]{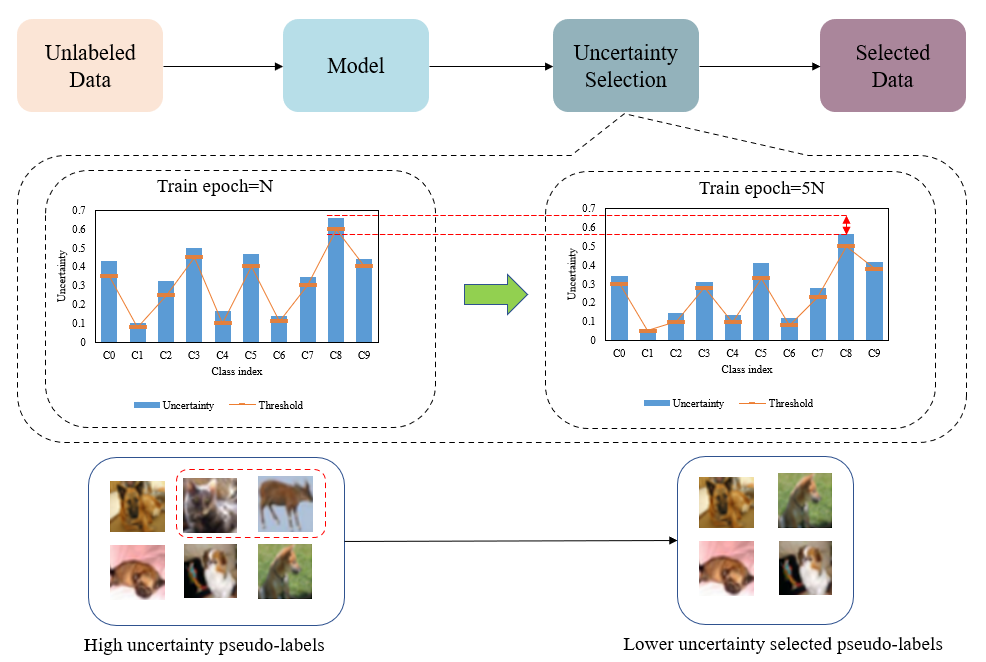}
    \caption{Evolution of uncertainty selection during training. Throughout various training stages, both the uncertainty of unlabeled data and the dynamic uncertainty threshold evolve over time. The proposed method prioritizes selecting images with lower uncertainty, enhancing model performance. In the figure, images outlined by red dashed rectangles indicate instances of high uncertainty that have been incorrectly classified.}
    \label{fig:uncertainty and uncertainty threhold}
\end{figure*}

%\begin{figure}[tp]
%    \centering
%    \includegraphics[width=8.5cm]{iccv_1.png}
%    \caption{Uncertainty selection changes as the training continues. The uncertainty of the unlabeled data and the dynamic uncertainty threshold are time-variant through different training stages. The proposed uncertainty selection selects images with lower uncertainty to improve the model performance. The images with red bounding boxes in the figure represent misclassified instances with high uncertainty.}
%    \label{fig:uncertainty and uncertainty threhold}
%\end{figure}
Semi-supervised learning, recognized for its ability to reduce the need for labeled data by leveraging abundant unlabeled data, often employs the generation of pseudo-labels \cite{lee2013pseudo} from model predictions for regularization training \cite{yun2019cutmix}. The efficacy of semi-supervised models is closely linked to the balance in distribution between labeled and unlabeled data \cite{yang2020rethinking}. In cases with long-tailed data, the skewness inherent in such datasets significantly impacts the quality of the generated pseudo-labels. This often leads to a disproportionate representation of dominant classes in pseudo-labels, negatively affecting the performance and robustness of models developed during the training process. 
%It is well known that semi-supervised learning reduces the cost of labeled data by utilizing large amounts of unlabeled data. These semi-supervised methods generate pseudo-labels\cite{lee2013pseudo} for unlabeled data from model predictions and use them for regularization training\cite{yun2019cutmix}. However, the model performance of semi-supervised learning depends on the degree distribution balance of labeled and unlabeled data \cite{yang2020rethinking}. When facing the long-tailed data, the deviation brought by the long-tailed data will have a greater impact on the quality of the generated pseudo-labels, which can lead to the pseudo-labels being more biased to most classes, and thus having a disastrous impact on the generated models in the process of training.

Uncertainty estimation\cite{gal2016dropout} reflects the dispersion degree of a random variable, and the uncertainty prediction of a model aids in assessing the reliability of the conditional probability distribution output by the model. Generally, higher model uncertainty correlates with less reliable predictions. Traditional semi-supervised learning methods generate pseudo-labels based solely on confidence, often filtering out unlabeled data that, despite meeting the confidence threshold, exhibit high uncertainty. This scenario leads to pseudo-labels that are seemingly confident yet unreliable, adversely affecting model performance. By leveraging uncertainty estimation to discern labels with both high confidence and high uncertainty, we can enhance the reliability of the filtered pseudo-labels, ultimately contributing positively to the training process of the model.

%Uncertainty estimation\cite{gal2016dropout} reflects the dispersion degree of a random variable, and the uncertainty prediction of a model can judge the reliability of the conditional probability distribution output by the model. The higher the uncertainty of the model, the less reliable the results it predicts. In the conventional semi-supervised learning method, the pseudo-labels are generated only by confidence, and the model screen out some unlabeled data with high uncertainty but the confidence meets the threshold. Although the data reach the confidence threshold, its high uncertainty indicates that it is not reliable. In this case, the label meets the confidence threshold, but it is not reliable. Such pseudo-labels will reduce model performance. If we can select those labels with high confidence but high uncertainty through uncertainty estimation of the samples, so as to make the filtered pseudo-labels more reliable, it will also be beneficial for the model training.

In the context of long-tailed data, the model encounters varying quantities of each data type, resulting in differing learning states for each category. Abundant head data leads to more comprehensive learning by the model, whereas sparse tail data often results in lower prediction confidence. Relying on a manually set or fixed threshold leads to two issues: 1) the predictive capability of the model varies throughout the training process, with early stages typically marked by high sample uncertainty; 2) the learning states of model for different data types within the long tail vary, and uncertainty estimation often reveals lower uncertainty for head data compared to tail data. Furthermore, threshold-based methods don't take into account how to select more reliable pseudo-labels for training. To address these challenges, we propose an uncertainty dynamic threshold approach, which more effectively selects reliable and diverse samples, catering to the unique learning requirements posed by long-tailed data distributions.

%Under the condition of long-tailed data, the model learns from different amounts of each data type, and the learning state of the model for each type is also different. More header data will make the model learn more information, on the contrary, the model cannot generate high confidence in the prediction of the tail type. If the threshold is set manually or is fixed, there are two consequences: 1. the prediction ability of the model is different in the whole training process. The uncertainty of the sample at the early stage is very high; 2. the model has different learning states for long-tailed data, and the uncertainty of the head is often lower than that of the tail when uncertainty estimation is carried out. Therefore, we propose an uncertainty dynamic threshold to select reliable and diverse samples more effectively.

To visually demonstrate the aforementioned challenges, the training process of the FixMatch method on CIFAR10-LT is demonstrated in Figure \ref{fig:uncertainty and uncertainty threhold}. We calculate the uncertainty of the unlabeled data by Monte Carlo dropout, and compute the uncertainty of all the classes separately when epoch=$N$ and epoch=$5N$, and obtain the mean value of the uncertainty according to each class. As shown in Figure \ref{fig:uncertainty and uncertainty threhold}, the model uncertainty decreases as training progresses, but it varies among different classes. Therefore, we propose an Uncertainty-Aware Dynamic Threshold Selection (UDTS) tailored for imbalanced semi-supervised learning. UDTS dynamically updates the uncertainty threshold based on the evolving ability of model to perceive different data classes at various training stages, thereby mitigating the impact of class imbalance on the network. This strategy effectively lowers the rate of pseudo-label misclassification, facilitating more effective learning from these labels by the network.

%Motivated by this problem, we do an experiment on CIFAR10-LT with FixMatch. We calculate the uncertainty of the unlabeled data by Monte Carlo dropout, and compute the uncertainty of all the classes separately when epoch=$N$ and epoch=$5N$, and obtain the mean value of the uncertainty according to each class. As shown in Figure \ref{fig:uncertainty and uncertainty threhold}, the uncertainty of the model decreases during the training stage, but the uncertainty of each type is different. Therefore, we propose uncertainty-aware dynamic threshold selection for imbalanced semi-supervised learning. According to the model perception ability for each data class at different training stages, the uncertainty threshold is dynamically updated to reduce the class-imbalanced influence on the network. This approach reduces the misclassification rate of pseudo-labels, enabling the network to better learn from the information contained in these pseudo-labels.

The main contributions of the current work are as follows: 

1) We propose an Uncertainty-Aware Dynamic Threshold Selection (UDTS) as a novel approach to tackle the challenge of long-tailed data distribution in semi-supervised learning. UDST dynamically adjusts selection thresholds for different classes, effectively adapting to the evolving proficiency of model in handling diverse data distributions. 

2) The feasibility and effectiveness of UDTS are theoretically underpinned and validated using Bayesian optimization and risk analysis. This theoretical derivation emphasizes the robustness and practical utility of UDTS in real-world scenarios.

3) We conducted extensive experiments on public datasets including CIFAR10/100-LT, STL-10-LT and TissueMNIST, validating the capability of UDTS in fostering more dynamic and accurate learning of long-tailed data traits, and mitigating overfitting in predominantly sampled classes.

%The main contributions of our work can be summarized as follows: 1. we propose an uncertainty-aware dynamic threshold selection so that the model can focus on reliable targets; 2. we propose an adaptive uncertainty threshold selection for different training stages and introduce uncertainty loss to reduce the model deviation; 3. we have achieved good performance on the public datasets CIFAR10/100-LT, STL-10-LT and TissueMNIST, demonstrating the effectiveness of our method.

\section{Related Work}
\noindent{{\bfseries Long-tailed recognition.}} In the real world, data often exhibits class imbalanced or long-tailed distribution. The solutions to this problem include data re-weighting \cite{ren2020not} or data re-sampling  \cite{ando2017deep}\cite{kim2020m2m}, which aim to balance the classes. Simple re-balancing based on class distribution makes the model overfitting in certain classes. Other methods include decoupling classifiers \cite{oh2022daso}\cite{kang2019decoupling}\cite{zhou2020bbn} and employing expert models for various classes, thereby recalibrating data distribution during loss computation. Different from the above methods, the current work focuses on correcting the bias in pseudo-label generation caused by long-tailed data in semi-supervised learning, which in turn affects model performance.

\noindent{{\bfseries Semi-supervised learning.} In the realm of semi-supervised learning, several approaches have been introduced in recent years to leverage unlabeled data. These include generating pseudo-labels based on model predictions \cite{lee2013pseudo}, and applying consistent regularization techniques \cite{sajjadi2016regularization}\cite{miyato2018virtual}. In addition, data augmentation strategies, exemplified by FixMatch \cite{sohn2020fixmatch} and ReMixMatch \cite{berthelot2019remixmatch}, employ advanced augmentation techniques such as Cutout \cite{berthelot2019remixmatch} and Random Augment \cite{zhang2017mixup}. When these approaches encounter long-tailed data, the tendency of model to bias predictions towards dominant classes can lead to a reduction in overall performance, as the pseudo-labels generated for unlabeled data are derived from these skewed model predictions.}

\begin{figure*}[htp]
    \centering
    \includegraphics[width=17.5cm]{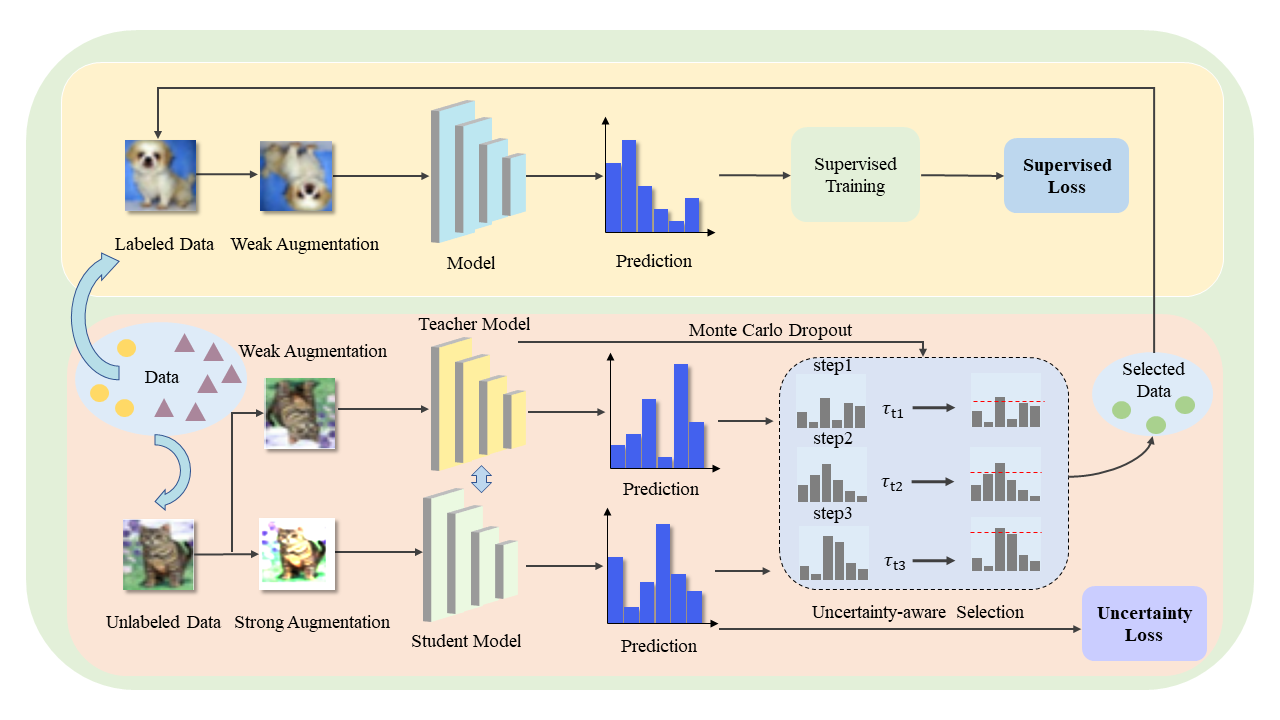}
    \caption{Overall framework of UDTS. ${{\tau }_{t1}}$, ${{\tau }_{t2}}$ and ${{\tau }_{t3}}$ are the dynamic uncertainty threshold in different steps. First, the input long-tailed data is divided into labeled and unlabeled data. A model is trained on the labeled data, and predictions are made on the unlabeled data. Additionally, the Monte Carlo Dropout method is employed to estimate the uncertainty of the predictions. After uncertainty-aware selection, relying on various learning states at different training stages, more reliable and diverse pseudo-labels are chosen using adaptive uncertainty thresholds. Simultaneously, the model is adjusted using uncertainty loss. The selected uncertain data and labeled data are combined and sent to the upper layers for supervised learning. This process is repeated in a loop until convergence.}
    \label{fig: Overall Framework}
\end{figure*}

\noindent{{\bfseries Semi-supervised learning with long-tailed data.} For semi-supervised learning, the pseudo-labels generated by itself produce certain deviations and have a certain influence on model prediction.  This issue becomes more pronounced in scenarios involving long-tailed data, where pseudo-labels are likely to manifest greater deviations, adversely affecting model performance. Some work deal with this problem through loss re-weighting \cite{blundell2015weight}, optimization \cite{kim2020distribution}, data re-sampling \cite{ando2017deep}\cite{kim2020m2m}, meta-learning \cite{pham2021meta}\cite{ren2020balanced}, ensemble learning\cite{xiang2020learning}\cite{laine2016temporal}.

\noindent{{\bfseries Uncertainty estimation and threshold selection.} The concept of uncertainty estimation in neural networks \cite{gal2016dropout}\cite{lakshminarayanan2017simple}\cite{blundell2015weight}\cite{louizos2016structured}\cite{welling2011bayesian} has been extensively explored, enhancing model robustness and reliability \cite{lakshminarayanan2017simple}\cite{malinin2018predictive}\cite{maddox2019simple}\cite{mukherjee2020uncertainty}. In medical image processing, the consistency regularization of uncertainty is used to improve segmentation accuracy \cite{xia20203d}\cite{yu2019uncertainty}. Uncertainty estimation is also used for model calibration\cite{xing2019distance}\cite{guo2017calibration}. There are few kinds of research on the sample selection of long-tailed data through uncertainty-aware. Confidence threshold selection strategies vary, ranging from manually setting fixed thresholds \cite{sohn2020fixmatch} to adaptive thresholding \cite{xu2021dash}\cite{lai2022smoothed}\cite{wang2022freematch} based on the training progression, also  including techniques like smooth adaptive weight adjustment \cite{lai2022smoothed}. While there is research on uncertainty in pseudo-labels \cite{rizve2021defense}, these often require fine-tuning multiple hyperparameters. UDTS diverges from these methods by employing a dynamic uncertainty threshold for model-based sample selection, thereby streamlining the process and enhancing the model adaptability to varying data distributions.

%For the selection of confidence threshold, there are also many kinds of selection through manual selection of different fixed thresholds \cite{sohn2020fixmatch} and adaptive selection of thresholds \cite{xu2021dash}\cite{lai2022smoothed}\cite{wang2022freematch} according to the training process, as well as the smooth adaptive weighting of weights \cite{lai2022smoothed}. At the same time, there is also a selection for the uncertainty of the pseudo-label \cite{rizve2021defense}, but the selection of uncertainty threshold requires multiple hyperparameters for debugging selection. Different from the above methods, our method estimates the uncertainty of the model using the dynamic uncertainty threshold to select the samples.

\section{Methods}
\noindent{\bfseries Framework of UDTS.} Figure \ref{fig: Overall Framework} illustrates our Uncertainty-Aware Dynamic Threshold Selection (UDTS) approach. Initially, the input long-tailed data undergo division into labeled and unlabeled segments, followed by network training to predict the unlabeled data. Concurrently, the estimated uncertainty of the predicted outcomes is determined through Monte Carlo Dropout. Subsequently, an uncertainty-aware selection process is employed, choosing more reliable and diverse pseudo-labels by adapting the uncertainty threshold to different learning states during various training stages. Additionally, the uncertainty loss helps fine-tune the model. The selected data featuring uncertainty, along with labeled data, are combined and forwarded to the upper level for supervised learning. This cyclic process continues iteratively until convergence. We illustrate the embedding of UDTS into the FixMatch method\cite{sohn2020fixmatch}.

%\noindent{\bfseries Framework of UDTS.} As shown in Figure \ref{fig: Overall Framework}, our method is built based on FixMatch\cite{sohn2020fixmatch} for convenience. First, the input long-tailed data are divided into labeled and unlabeled data, and the model is obtained after network training for the prediction of unlabeled data. At the same time, the uncertainty of the predicted results is estimated using Monte Carlo Dropout. After uncertainty-aware selection, more reliable and diversified pseudo-labels are selected using an adaptive uncertainty threshold according to each kind of learning state at different training stages. At the same time, uncertainty loss is used to adjust the model. Data selected with uncertainty and labeled data are combined and sent to the upper level for supervised learning. Then the model is cycled until convergence.

\subsection{Problem Setting}
We assume that the dataset $D=\left\{ \left( {{x}_{i}},{{y}_{i}} \right) \right\}_{\text{i}=1}^{N}$ is divided into a labeled dataset ${{D}_{lb}}=\left\{ \left( x_{i}^{l},y_{i}^{l} \right) \right\}_{\text{i}=1}^{m}$, and an unlabeled dataset ${{D}_{ulb}}=\left\{ \left( x_{i}^{u},y_{i}^{u} \right) \right\}_{\text{i}=1}^{n}$ ($N=m+n$, $m\ll n$), where each label corresponds to an image classification. The imbalance ratio of labeled and unlabeled data is ${{\gamma }_{lb}}$ and ${{\gamma }_{ulb}}$. The labeled and unlabeled data are arranged in descending order, ${{m}_{1}}>{{m}_{2}}>{{m}_{3}}>\ldots >{{m}_{C}}$, ${{n}_{1}}>{{n}_{2}}>{{n}_{3}}>\ldots >{{n}_{C}}$. The data of each class are distributed from more to less according to the long-tailed data. Therefore, the imbalance ratio ${{\gamma }_{lb}}$ is defined as ${{\gamma }_{lb}}=\frac{{{m}_{1}}}{{{m}_{C}}}$, where ${{m}_{1}}$ and ${{m}_{c}}$ are head and tail class, respectively. Similarly, ${{\gamma }_{ulb}}$ is defined as ${{\gamma }_{\text{u}lb}}=\frac{{{n}_{1}}}{{{n}_{C}}}$. When the long-tailed data $D$ is fed into the model, the model learns the labeled data ${{D}_{lb}}$, and the unlabeled data is generated by the network prediction directly. The $p_{C}^{\left( i \right)}$ probability represents the probability that the $i$-th sample is predicted to belong to a certain class $C$. If the $p_{C}^{\left( i \right)}$ probability is greater than the threshold set $\tau $, the model assigns the unlabeled data a pseudo-label of class $C$.

\begin{equation}
{y_{C}^{\left( i \right)}=\mathbbm{1}\left[ p_{C}^{\left( i \right)}\ge \tau  \right]}
\end{equation}
\noindent{where $\tau$ denotes a threshold utilized for generating pseudo-labels.}

\begin{figure*}[tp]
    \centering
    \includegraphics[width=16cm]{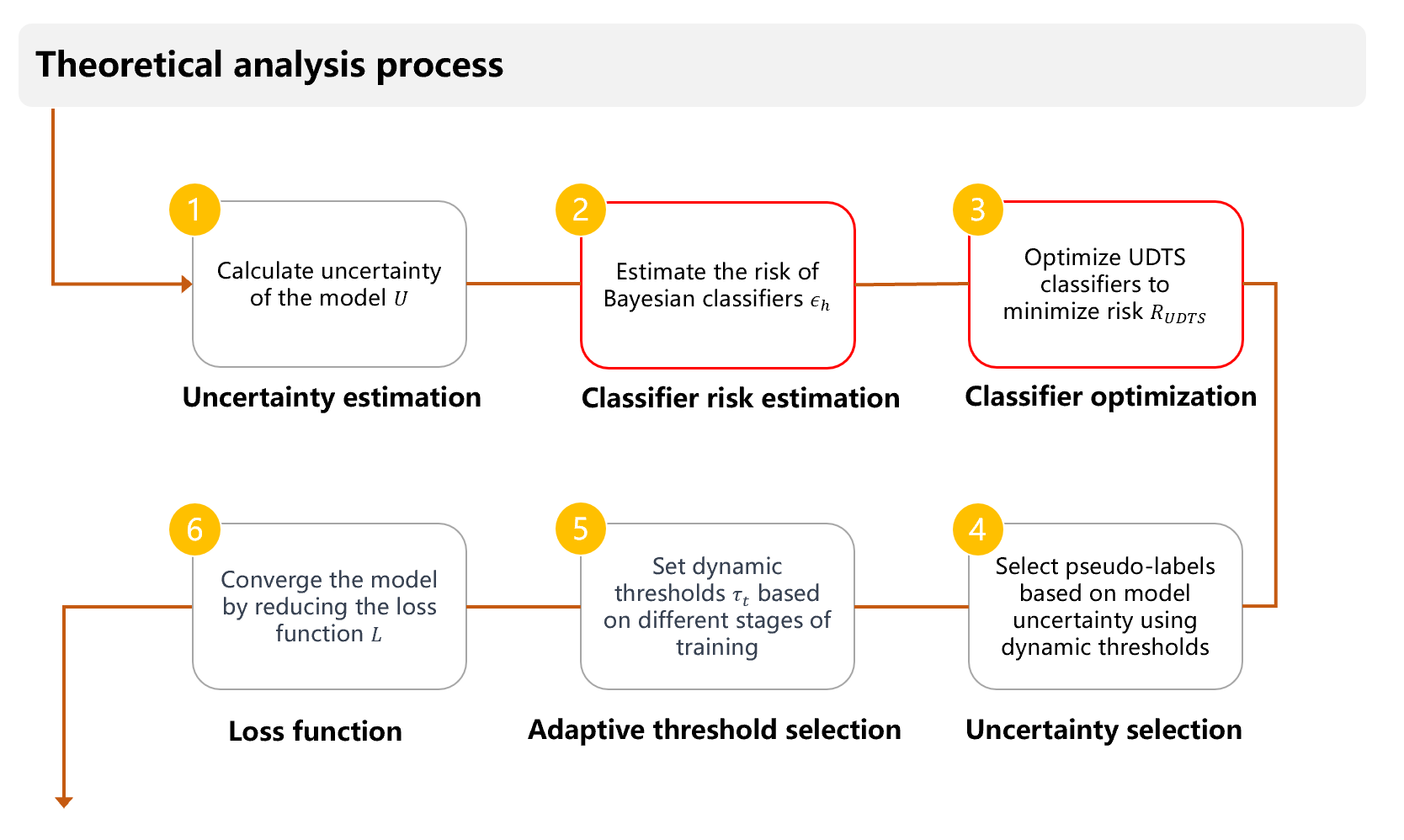}
    \caption{Theoretical analysis process of UDTS. The red boxes highlight the primary innovative contributions. Firstly, model uncertainty is computed, followed by estimating the risk of the Bayesian classifier. Subsequently, optimizing this classifier occurs. Dynamic thresholds are then designed, and the selection is based on the computed uncertainty. Finally, a loss function is designed to converge the model.}
    \label{fig:theoretical analysis}
\end{figure*}

\subsection{Theoretical Analysis}
The theoretical analysis process of UDTS is depicted in Figure \ref{fig:theoretical analysis}, with each component's theoretical analysis corresponding to subsequent textual sections. Regarding the classification task, the definition of the class center is as follows: In a classification task with $C$ categories, each category comprises $N_i$ samples. Denoting $x_{j}^{i}$ as the $j$-th sample within category $i$, the center of the $i$-th category is represented as:
%The theoretical analysis process of UDTS is depicted in Figure \ref{fig:theoretical analysis}, where each component's theoretical analysis corresponds to subsequent textual sections. In the context of a classification task, we define the class center as follows: In a classification task with $C$ categories, each category has $N_i$ samples. Let $x_{j}^{i}$ be the $j$-th sample in category $i$. The center of the $i$-th category can be represented as:

\begin{equation}
    m_{i} = \frac{1}{N}{\sum_{j = 1}^{N}x_{j}^{i}}
\end{equation}

Therefore, the intra-class distance for the $i$-th category can be defined as:

\begin{equation}
s_{i} = \frac{1}{N}{\sum_{j = 1}^{N}\left| \middle| x_{j}^{i} - m_{i} \middle| \right|^{2}}
\end{equation}

\noindent{where $
\left| \middle| ~~ \middle| \right|^{2}$ represents the $L2$ norm, representing the Euclidean distance and yielding a $C$-dimensional intra-class distance vector. Within this vector, $s_i$ denotes the intra-class distance within the $i$-th category.}

For multi-class problems, assuming there are $C$ categories, the decision boundaries for each category can be represented as:

\begin{equation}
w_{i}^{T}x + b_{i} = 0,~i = 1,\ldots,~C
\end{equation}

\noindent{where $w$ is the weight vector, $b$ is the bias term, $x$ is the input feature vector, $w_i$ represents the weight vector for the $i$-th category, and $b_i$ stands for the bias term for the $i$-th category. When a sample $x$ satisfies $w_{i}^{T}x + b_{i} > 0$, it is classified as belonging to the $i$-th category.}

During the training process, the impact of long-tailed data often reduces the distances between misclassified categories and majority classes, causing minority class data to be incorrectly absorbed into the majority classes. This is an erroneous phenomenon. To counteract this issue, we introduce an uncertainty measure. If the model assigns data with high uncertainty to a single class during classification, we remove such data instances, ensuring that highly uncertain data does not influence the determination of class centers.

Uncertainty can be categorized into aleatoric uncertainty and epistemic uncertainty. Aleatoric uncertainty refers to inherent noise within the data, which is unavoidable. On the contrary, epistemic uncertainty pertains to the uncertainty linked to the deep learning model itself. Inaccurate model predictions may stem from suboptimal training, insufficient or imbalanced training data. Epistemic uncertainty is based on the estimation of model parameter uncertainty during the training process, and it can be approximately estimated and mitigated.

\begin{equation}
u = ~u_{A} + u_{E}
\end{equation}

$u_A$ denotes aleatoric uncertainty, which is related to inherent or random uncertainty.
$u_E$ stands for epistemic uncertainty, linked to model uncertainty.

As aleatoric uncertainty is essentially a constant that cannot be avoided, we primarily focus on epistemic uncertainty, denoted as $u$, which signifies the model uncertainty.

Based on the previous research \cite{gal2016dropout}\cite{gneiting2007strictly}\cite{lakshminarayanan2017simple}, deep ensemble networks allow us to estimate the model uncertainty.

\begin{equation}
p\left( y \middle| x \right) = M^{- 1}{\sum_{m = 1}^{M}{P_{\theta_{m}}\left( y \middle| x,\theta_{m} \right)}}
\end{equation}

\noindent{where $M$ represents the number of neural networks, and $\theta_m $ represents the parameters.}

In the current work, we attempt model ensembling as a means to quantify uncertainty. Experimental results reveal that setting the dropout rate to 0.5 and performing 10 forward passes for predictions from the model ensemble, with the standard deviation serving as the measure of uncertainty, yields the most accurate results. 

% \begin{equation}
% U=\frac{1}{N}\sum_{i=1}^Nu_i
% \end{equation}

% \begin{equation}
% \sigma=\sqrt{\frac{1}{N}\sum_{i=1}^N(X_i-\mu)^2}
% \end{equation}

\begin{equation}
u = \sigma\left( p\left( y \middle| x \right) \right)
\end{equation}

\noindent{where $\sigma$ stands for standard deviation. We define the impact factor of uncertainty, denoted as $\varepsilon$, on model predictions as $\left. ~\varepsilon \right.\sim\frac{N_{t}}{N_{h}}$, where $0 < \varepsilon < 1$.}

When making inferences on long-tailed data, the risk associated with the head-class data is defined as follows:

\begin{equation}
\small
\epsilon_{h} = E\left\lbrack {y_{h} \neq h\left( z_{h} \right)} \middle| {z_{h} = R_{b \in h}\left( {(1 - \epsilon)f_{\theta_{m}}\left( n_{h} \right) + f_{\theta_{m}}\left( n_{t} \right)} \right)} \right\rbrack
\end{equation}

\noindent{where $Z$ represents the features, $R$ is the feature function, and $y_{h} \neq h\left( z_{h} \right)$ represents data in the classification where the actual label is not the head-class category.}

In this paper, we define our model as follows:

\begin{equation}
f_{UDTS}(x) = ~f(x) - \epsilon_{h} - \epsilon_{t}
\end{equation}

\noindent{where $\epsilon_{t}$ represents the risk of the training data distribution. As the training distribution becomes more imbalanced, it leads to increased risk for the model.}

For long-tailed data distributions, we minimize the objective function during training to mitigate the aforementioned uncertainty risk, thereby minimizing the misclassification rate of model.

\begin{equation}
\small
min\frac{1}{n}{\sum\limits_{i = 1}^{n}{L\left( {f_{\theta}\left( x_{i} \right),~y_{i}} \right)}} = min\left( \frac{1}{N}{\sum\limits_{i = 1}^{N}\left. L\left( {f_{\theta}\left( x_{i} \right),~y_{i}} \right) + \epsilon_{h} \right)} \right).
\end{equation}
\begin{equation}
\small
= min\left( \frac{1}{N}{\sum\limits_{i = 1}^{N}{L\left( {f_{\theta}\left( x_{i} \right),~y_{i}} \right)}} + \lambda\frac{1}{N}{\sum_{i = 1}^{N}{\sum_{k = 1}^{K}{w_{k}y_{ik}{\log\left( p_{ik} \right)}}}} \right)
\end{equation}

In the context of cross-entropy loss function $L\left( {f_{\theta}\left( x_{i} \right),~y_{i}} \right)$, where $N$ represents the number of samples, and $\lambda$ is a weight.

The classifier reaches optimality by minimizing the loss function associated with uncertainty risk in the target distribution, thereby minimizing the overall risk. Consequently, the obtained classifier is optimal.

\begin{equation}
R_{t}\left( f_{UDTS} \right) \leq R_{t}\left( f_{others} \right)
\end{equation}
\noindent{where $R_{t}\left( f_{UDTS} \right)$ represents the overall risk of the model, and $R_{t}\left( f_{others} \right)$ represents the risk of other models. Following our uncertainty-based selection, the risk of our model is either less than or equal to the risk of other models. In other words, our model can achieve a lower misclassification rate compared to other models.}

Based on the estimated uncertainty of the data during the training phase and the distribution of predictions made by the model, we strategically select more dependable labels. We establish a scoring function aimed at minimizing $S_\theta$, allowing the model to learn more reliable and accurate predictions. This scoring function relates to the model uncertainty in predicting data $u_{y_{h} \neq h{(z_{h})}}$ and the prediction distribution $p_{\theta}\left( y \middle| x \right)$. By separating or filtering out tail-class data that the model tends to classify as head-class data based on uncertainty measurements, the model performance is improved.
%Based on the estimated uncertainty of the data during the training process and the model's prediction distribution, we select more reliable labels. We set up a scoring function to minimize $S_\theta$, allowing the model to learn more reliable and accurate predictions. This scoring function is related to the model's uncertainty in predicting data $u_{y_{h} \neq h{(z_{h})}}$ and the prediction distribution $p_{\theta}\left( y \middle| x \right)$. By separating or filtering out tail-class data that the model tends to classify as head-class data based on uncertainty measurements, the model performance is improved.

Therefore, we incorporate uncertainty measurement as a criterion during model training. By excluding data with high uncertainty from the training process, we mitigate the dominance of majority classes and diminish the effect of uncertainty on the model. This, in turn, enhances the classification performance of model on long-tailed data.

\begin{equation}
L\left( {y,p} \right) = - \frac{1}{N}{\sum_{i = 1}^{N}{\sum_{k = 1}^{K}{w_{k}y_{ik}log\left( p_{ik} \right)}}}
\end{equation}

\noindent{where, $p_ik$ represents the predicted probability of model that the $i$-th sample belongs to the $k$-th class, $y_ik$ represents the true class label for the $i$-th sample, with $y_ik=1$ denoting membership in the $k$-th class, and $y_ik=0$ indicating otherwise. In addition, $w_k$ represents the weight for the $k$-th class.}

\subsection{Uncertainty-aware Selection }
The semi-supervised learning method encounters a significant challenge in generating pseudo-labels, particularly in scenarios involving long-tailed data. This arises from the substantial reliance on pseudo-labels. In the process of generating these labels, most models use the SoftMax function to gauge the confidence probability of a class and then designate the class with the highest confidence as the pseudo-label. Relying solely on confidence often leads to the selection of incorrect predictions with high confidence scores during model training. To tackle this issue, we introduce an uncertainty-guidance module to enhance the accuracy of pseudo-label selection.

Incorporating Monte Carlo dropout \cite{gal2016dropout} into the network enables uncertainty estimation, which encompasses both the model confidence and its prediction uncertainty. As the model prediction uncertainty increases, so does the prediction error rate. Therefore, we can mitigate potential pseudo-label inaccuracies by considering prediction uncertainty, enhancing the quality of imbalanced pseudo-label data generated by the model when trained on long-tailed data. This enables the model to progressively learn from more reliable sources, thereby enhancing the accuracy of pseudo-labels.

Given a batch of images, the model not only generates the prediction of the target but also estimates the uncertainty of each target. With the guidance of uncertainty, the model is optimized to prioritize more reliable targets.

We estimate uncertainty using Monte Carlo dropout and perform $T$ random forward passes for each input, enhancing robustness by adding random noise to the model. After acquiring the uncertainty for each batch of images, we proceed with uncertainty selection. Specifically, we perform $T$ iterations of random forward transmission for each input teacher model with random dropout and input Gaussian noise.

\begin{equation}
{{\mu }_{C}}=\frac{1}{T}\underset{t}{\mathop \sum }\,p_{t}^{C}
\end{equation}

\begin{equation}
u=-\underset{c}{\mathop \sum }\,{{\mu }_{C}}\log {{\mu }_{C}}
\end{equation}
\noindent{where $p_{t}^{C}$ is the probability of class $C$ in the $t$ prediction.}

Although pseudo-labels are a general method and are mode-independent, the performance of semi-supervised learning method based on pseudo-labels tends to suffer when the input of pseudo-labels is long-tailed data. This occurs because the model has limited exposure to tail data, leading to significant bias toward the majority classes during initial predictions. Consequently, a substantial number of incorrect pseudo-labels are generated during training, resulting in subpar model predictions. We introduce uncertainty guidance to assess the uncertainty associated with model-generated pseudo-labels, allowing us to select and calibrate pseudo-labels affected by class imbalance. Given the inherent characteristics of long-tailed data distributions, predictions often exhibit more substantial deviations, leading to higher uncertainty levels in these pseudo-labels. We mitigate this by implementing a mechanism for the selection and removal of high-uncertainty pseudo-labels.

We utilize two metrics, model confidence and model uncertainty, for evaluation purposes. The initially trained model is applied to the unlabeled data, through the previously configured Monte Carlo dropout and SoftMax layer of the network, various types of confidence and uncertainty in data predictions are ultimately derived and synthesized for the model. Subsequently, the threshold value is used for selection. We calculate weights for surplus uncertainty and confidence independently and then proceed to select samples with the highest overall score, which corresponds to samples exhibiting low uncertainty and high confidence. These selected samples are then forwarded for subsequent network training. The selection of the uncertainty dynamic threshold is discussed in the next subsection.

\renewcommand\arraystretch{2}
\begin{algorithm}[t]
\algsetup{linenosize=\large}
\normalsize
\caption{\normalsize{Pseudo code of UDTS}}
\label{alg:alg1}
\begin{algorithmic}
% \STATE {\textbf{Input}: A video sequence with  \emph{N} frames}
% \STATE {\textbf{Input}: A video sequence with  \emph{N} frames}
\STATE {\textbf{Input}: a set of labeled data, ${{D}_{lb}}$, and a set of unlabeled data, ${{D}_{ulb}}$,}
\STATE {\textbf{Output}: a trained model ${{f}_{\theta}}$}
\STATE {\textbf{Parameters}: $\theta$ (parameters of Wide-ResNet-28-2 and our method)}
\STATE \textbf{for} $i$ in range (epochs) \textbf{do}
\STATE \hspace{0.25cm} 1: Train a model ${{f}_{{\theta}_{i-1}}}$
\STATE \hspace{0.25cm} 2: Use ${{f}_{{\theta}_{i-1}}}$ to ${{D}_{ulb}}$
\STATE \hspace{0.25cm} 3: Calculate the uncertainty $u\left( p_{c}^{\left( i \right)} \right)$ of each ${{D}_{ulb}}$ through 
\STATE \hspace{0.25cm} Monte Carlo dropout (Equation \ref{111})
\STATE \hspace{0.25cm} 4: Compare ${{\tau_t(c)}}$ with $u\left( p_{c}^{\left( i \right)} \right)$ (Equation \ref{calculate1}, \ref{calculate2})
\STATE \hspace{0.25cm} 5: Select reliable data ${{D}_{select}}$
\STATE \hspace{0.25cm} 6: Combine ${{D}_{lb}}$ with ${{D}_{select}}$
\STATE \hspace{0.25cm} 7: Update ${\tau_t(c)}$
\STATE \hspace{0.2cm} 8: Calculate loss and update loss (Equation \ref{loss1}, \ref{loss2})
\STATE \hspace{0.25cm} 9: Update ${{f}_{\theta}}$

\end{algorithmic}
\label{alg2}
\end{algorithm}

\subsection{Adaptive Threshold Selection}
Algorithm \ref{alg:alg1} shows the pseudo-code algorithm, the core part of the algorithm consists of the following steps within UDTS. Since the distribution of long-tailed data is not as balanced as the distribution of existing datasets, the difficulty of learning each class is different when the model learns long-tailed data. Moreover, during the early stages of training, the model tends to learn from most data classes, leading to a natural prediction bias toward these majority classes. Consequently, the choice of threshold is crucial. In the early training phase, adopting a lower threshold is preferred, as it aligns with the model inclination toward most classes. As training progresses, the threshold will slowly increase. In the middle stage of training, to mitigate potential pseudo-label deviations, we adjust the threshold upwards to reduce such deviations. In the later stage of training, because the data of the tail class will be more difficult to determine, a specific threshold value is adopted for uncertain selection for the tail class that is more difficult to predict.

\begin{figure*}[htp]
    \centering
    \includegraphics[width=18cm]{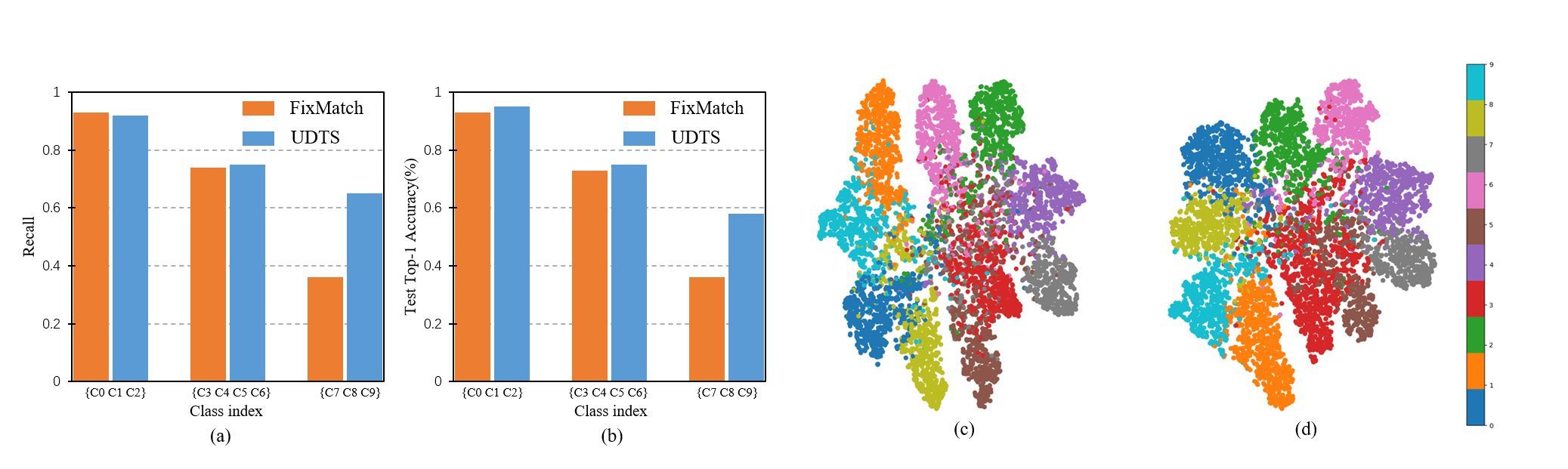}
    \caption{Analysis of FixMatch and our method in terms of recall, test accuracy, and t-SNE on CIFAR10-LT. Figure (a) and Figure (b) show that the class indexes of the X-axis are sorted by the class size, with C0 as the head class and C9 as the tail class. Figure (c) and Figure (d) show the t-SNE of FixMatch and UDTS respectively. UDTS gets higher recall and test accuracy compared with FixMatch.}
    \label{fig: t-SNE}
\end{figure*}

\begin{figure*}[htp]
    \centering
    \includegraphics[width=18cm]{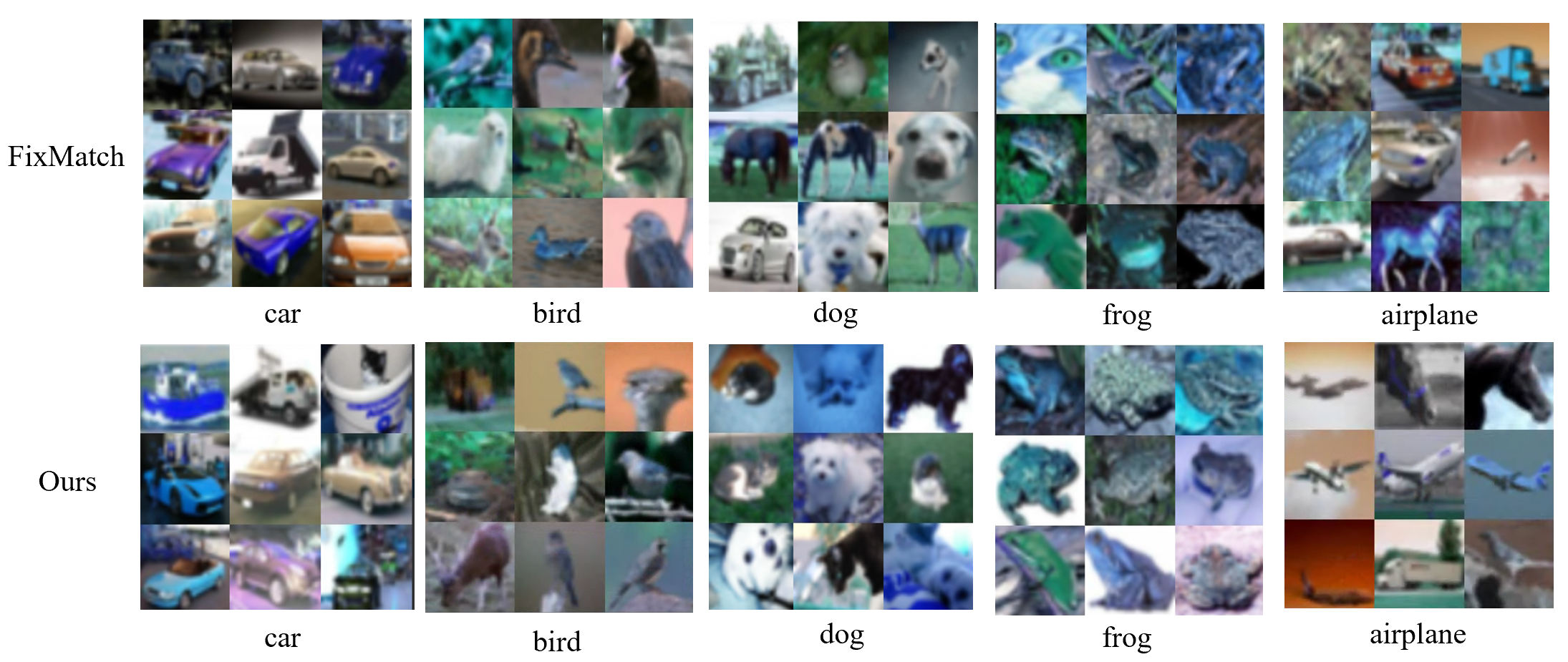}
    \caption{Visualization presenting the results on the CIFAR-10 dataset under the same experimental setup. From left to right, the categories are arranged from head class to tail class, with the most abundant category being \textquotedblleft car\textquotedblright and the least abundant being \textquotedblleft airplane\textquotedblright.}
    \label{fig: visualization}
\end{figure*}

Taking into account the relationship between the confidence and uncertainty of pseudo-labels generated by the model, we set a threshold of uncertainty of the predicted samples to select the correct pseudo-labels, and balance the number of various samples according to the degree of class-imbalanced. By selecting different uncertainty thresholds for long-tailed data and screening the uncertainty of unlabeled data, we can obtain samples with high confidence and low uncertainty, which will be more reliable. We combine the pseudo-label ${{D}_{s}}$ after uncertainty selection with the labeled dataset ${{D}_{lb}}$ and send it into the model for training.

Due to the distribution characteristics of long-tailed data, the exposure of model to samples from various classes varies, with the head classes receiving more attention than the tail ones during the learning process. When it comes to making predictions, using a fixed threshold can lead to predictions that are biased towards the majority classes, hampering the model training. Hence, we propose a dynamic threshold strategy that tailors threshold values to individual classes based on their specific learning complexities within the long-tailed dataset. This approach enables the model to make more accurate and diverse predictions, enhancing its overall performance. 

Our initial threshold design employs a low fixed threshold, which gradually increases as the number of training iterations progresses. Because predicting a large amount of unlabeled data after each model training round consumes a significant amount of time. Following the approach in FreeMatch \cite{wang2022freematch}, we use Exponential Moving Average (EMA) as a confidence estimate to save time required for predictions.

The initial threshold is set as follows considering the imbalance of each type of data.

\begin{equation}
{{\tau }_{t}}=\left\{ \begin{array}{*{35}{l}}
   {\gamma }_{C} \frac{1}{C}, & \text{ if }t=0,  \\
   \lambda {{\tau }_{t-1}}+(1-\lambda )\frac{1}{\mu B}\sum\limits_{b=1}^{\mu B}{\max }({{q}_{b}}), & \text{ otherwise, }  \\
\end{array} \right.
\label{111}
\end{equation}

\noindent{where ${{\gamma }_{C}}$ is the imbalance degree of class $C$ data, which is the imbalance ratio between class $C$ and head data of long-tailed distribution data. The imbalance ratio of class $C$ to tail is denoted as ${{\gamma }_{imb}}$, and $\lambda$ represents the EMA coefficient.}

\begin{equation}
\label{eq-salt}
\tilde{p}_t(c) = \begin{cases}
 {\gamma }_{c}\frac{1}{C}, & \text{ if } t=0, \\
 \lambda \tilde{p}_{t-1}(c) + (1-\lambda) \frac{1}{\mu B}\sum_{b=1}^{\mu B} q_b(c), & \text{ otherwise, }
\end{cases}
\end{equation}

We estimate the learning state of each class under the long-tailed data by calculating the expectation of all classes predicted to be class $C$. The weight of the learning state is adjusted based on the degree of imbalance in the long-tailed data. 

\begin{equation}
        \tau_t(c) =  \operatorname{MaxNorm}(\tilde{p}_t(c)) \cdot \tau_t 
        % = \frac{\tilde{p}_{t}(c)}{\max\{\tilde{p}_{t}(c) : c \in [C]\}} \cdot \tau_t
\end{equation}
\begin{equation}
        u_t(c) =  \operatorname{MaxNorm}(\tilde{u}_t(c))
        % = \frac{\tilde{u}_{t}(c)}{\max\{\tilde{u}_{t}(c) : c \in [C]\}}
\end{equation}

We predict the sample uncertainty of each class and summarize it to get $p$. The adaptive threshold value ${{\tau }_{t}}$ and $p$ with maximum normalization can be obtained.

\begin{equation}
\tau_t = \lambda \tau_{t-1} + (1-\lambda) \frac{1}{\mu B} \sum_{b=1}^{\mu B} max (q_b) 
\end{equation}

Afterward, the normalized uncertainty is compared to the respective threshold value. $u\left( p_{c}^{\left( i \right)} \right)$ is the $i$-th data predicted to be class $C$ uncertainty. When the uncertainty value is below the adaptive threshold ${{\tau }_{t}}$, it means that the sample possesses sufficient certainty to be selected.
\begin{equation}
{{\theta }_{c}}=\mathbbm{1}\left[ u\left( p_{c}^{\left( i \right)} \right)\le {{\tau }_{t}} \right]\
\label{calculate1}
\end{equation}

When the confidence exceeds the original threshold value, ${{\theta }_{c}}=1$, indicating that the sample exhibits  higher confidence and reliability. It is then selected and added to ${{D}_{s}}$.
\begin{equation}
{{\theta }_{c}}=\mathbbm{1}\left[ u\left( p_{c}^{\left( i \right)} \right)\le {{\tau }_{t}}\left]  \mathbbm{1}\right[p_{c}^{\left( i \right)}\ge {{\tau }_{c}} \right]
\label{calculate2}
\end{equation}

First, the network trains the model on the label data ${{D}_{lb}}$. Then, by comparing the adaptive uncertainty threshold and confidence, we can screen out more reliable pseudo-labels. These pseudo-labels are incorporated into the labeled dataset, and the network is reinitialized for training.

\renewcommand\arraystretch{1.3}
\begin{table*}
\normalsize    
\centering
\caption{Accuracy comparison with other methods on CIFAR10-LT and CIFAR100-LT.}
\vspace{0.75\baselineskip} 
\label{table1}
\setlength{\tabcolsep}{18pt}
%\resizebox{\textwidth}{45mm}
\begin{tabular}{lcccc}\hline
\multirow{7}*{Algorithm} & \multicolumn{2}{c}{CIFAR-10-LT} & \multicolumn{2}{c}{CIFAR-100-LT} \\  
     &${{\gamma }_{lb}}$=${{\gamma }_{ulb}}$=150 & ${{\gamma }_{lb}}$=${{\gamma }_{ulb}}$=100  & ${{\gamma }_{lb}}$=${{\gamma }_{ulb}}$=10 & ${{\gamma }_{lb}}$=${{\gamma }_{ulb}}$=15  \\ 
     \cmidrule(lr){2-2} \cmidrule(lr){3-3} \cmidrule(lr){4-4} \cmidrule(lr){5-5}
     &$N$ = 1500 & $N$ = 500& $N$ = 150 & $N$ = 150 \\
     &$M$ = 3000 & $M$ = 4000 & $M$ = 300 &$M$ = 300  \\
     \cmidrule(lr){1-1}\cmidrule(lr){2-3} \cmidrule(lr){4-5} 
Supervised   & 59.79${\pm}${\fontsize{10pt}{14pt}\selectfont 0.5} & 46.63${\pm}${\fontsize{10pt}{14pt}\selectfont 0.88}  & 48.26${\pm}${\fontsize{10pt}{14pt}\selectfont 0.19} & 45.69${\pm}${\fontsize{10pt}{14pt}\selectfont 0.26}  \\
FixMatch\cite{sohn2020fixmatch}    & 73.01${\pm}${\fontsize{10pt}{14pt}\selectfont 0.57} & 72.41${\pm}${\fontsize{10pt}{14pt}\selectfont 1.71}   & 57.76${\pm}${\fontsize{10pt}{14pt}\selectfont 0.6} & 54.29${\pm}${\fontsize{10pt}{14pt}\selectfont 0.5}  \\
w/ CReST\cite{wei2021crest}     & 74.47${\pm}${\fontsize{10pt}{14pt}\selectfont 0.39} & 74.21${\pm}${\fontsize{10pt}{14pt}\selectfont 0.76}  & 57.92${\pm}${\fontsize{10pt}{14pt}\selectfont 0.4} & 53.48${\pm}${\fontsize{10pt}{14pt}\selectfont 1.25}  \\
w/ CReST+\cite{wei2021crest}        & 74.59${\pm}${\fontsize{10pt}{14pt}\selectfont 0.66} & 76.38${\pm}${\fontsize{10pt}{14pt}\selectfont 1.37}  &  58.13${\pm}${\fontsize{10pt}{14pt}\selectfont 0.23} & 54.65${\pm}${\fontsize{10pt}{14pt}\selectfont 0.39} \\
w/ DARP\cite{kim2020distribution}         & 74.73${\pm}${\fontsize{10pt}{14pt}\selectfont 0.3} & 74.67${\pm}${\fontsize{10pt}{14pt}\selectfont 0.76}   & 58.22${\pm}${\fontsize{10pt}{14pt}\selectfont 0.23} & 54.89${\pm}${\fontsize{10pt}{14pt}\selectfont 0.42} \\
% w/ ABC\cite{lee2021abc}         & 81.73${\pm}${\fontsize{10pt}{14pt}\selectfont 0.25} & 82.02${\pm}${\fontsize{10pt}{14pt}\selectfont 0.58}   & 59.71${\pm}${\fontsize{10pt}{14pt}\selectfont 0.22} & 56.77${\pm}${\fontsize{10pt}{14pt}\selectfont 0.35}  \\
w/ DASO\cite{oh2022daso}         & 71.97${\pm}${\fontsize{10pt}{14pt}\selectfont 0.51} & 68.62 ${\pm}${\fontsize{10pt}{14pt}\selectfont 0.67}  & 59.01${\pm}${\fontsize{10pt}{14pt}\selectfont 0.16} & 55.75${\pm}${\fontsize{10pt}{14pt}\selectfont 0.29}  \\
w/ SAW\cite{lai2022smoothed}         & 76.75${\pm}${\fontsize{10pt}{14pt}\selectfont 0.23} & \textbf{77.73} ${\pm}${\fontsize{10pt}{14pt}\selectfont 0.81}  & 58.62${\pm}${\fontsize{10pt}{14pt}\selectfont 0.29} & 55.52${\pm}${\fontsize{10pt}{14pt}\selectfont 0.23}  \\
w/ DASH\cite{xu2021dash}         & 73.54${\pm}${\fontsize{10pt}{14pt}\selectfont 0.89} & 75.87 ${\pm}${\fontsize{10pt}{14pt}\selectfont 0.62}  & 58.37${\pm}${\fontsize{10pt}{14pt}\selectfont 0.26} & 54.5${\pm}${\fontsize{10pt}{14pt}\selectfont 0.27}  \\
w/ Debiaspl\cite{wang2022debiased}         & 73.41${\pm}${\fontsize{10pt}{14pt}\selectfont 0.47} & 73.49${\pm}${\fontsize{10pt}{14pt}\selectfont 1.05}   & 58.01${\pm}${\fontsize{10pt}{14pt}\selectfont 0.32} & 54.54${\pm}${\fontsize{10pt}{14pt}\selectfont 0.24}  \\
w/ UDTS (ours)     & \textbf{77.44}${\pm}${\fontsize{10pt}{14pt}\selectfont 0.73} & 76.48${\pm}${\fontsize{10pt}{14pt}\selectfont 1.65}  & \textbf{59.82}${\pm}${\fontsize{10pt}{14pt}\selectfont 0.23} & \textbf{56.28}${\pm}${\fontsize{10pt}{14pt}\selectfont 0.24}  \\ \hline
\multicolumn{5}{c}{} 
\end{tabular}
\label{tab_cifar}
\end{table*}

\subsection{Loss Function}

Under the long-tailed data, the pseudo-labels generated by the trained semi-supervised network model exhibit significant class imbalance. These pseudo-labels can introduce a substantial bias, often favoring most classes or a specific class. Incorrectly assigned pseudo-labels lead to the inclusion of mislabeled data during training, greatly impacting the model outcomes and exacerbating model deviation. To mitigate this issue, we employ a multi-class cross-entropy loss \ref{loss1}.

\begin{equation}
{{\mathcal{L}_{CE}}}=\frac{1}{{{N}_{l}}}\underset{i=1}{\overset{{{N}_{l}}}{\mathop \sum }}\,CE\left( {{y}_{i}},\widehat{{{y}_{i}}} \right)+\frac{\lambda }{{{N}_{u}}}\underset{i=1}{\overset{{{N}_{u}}}{\mathop \sum }}\,{{\omega }_{i}}CE\left( {{y}_{i}},\widehat{{{y}_{i}}} \right)
\label{loss1}
\end{equation}
\noindent{where, $N_l$ and $N_u$ represent the amount of labeled data and unlabeled data respectively; $\mathrm{CE}$ represents the cross entropy loss function; $y_i$ represents the actual label of sample $i$, $\hat{y}_i$ represents the predicted label of sample $i$; $w_i$ represents the weight of sample $i$, and $\hat{y}_i$ represents the weight of sample $i$; $\lambda$ indicates the parameter for weight adjustment.}

For unlabeled data, we can reduce the deviation of pseudo-labels to the model by calculating ${\theta }_{c}$ of sample uncertainty selection beforehand. If ${\theta }_{c}=0$, the unlabeled data will not be considered in the loss function computation. This results in the utilization of more reliable and high-confidence pseudo-labels while filtering out some noise during training. In comparison to traditional pseudo-labeling methods, our approach enhances model performance. By using the computed uncertainty $u$, we filter out unreliable (high uncertainty) samples and select more reliable targets for the model to learn. This leads to the formulation of the following uncertainty loss function \ref{loss2}.

\begin{equation}
{{\mathcal{L}_{u}}}=-\frac{1}{B}\underset{j=1}{\overset{M}{\mathop \sum }}\,\underset{i=1}{\overset{N}{\mathop \sum }}\,\theta _{c}^{\left( i \right)}{{p}_{ij}}log{{q}_{ij}}
\label{loss2}
\end{equation}

\noindent{where ${B}$ is the batch of class $C$ data, $M$ denotes the number of classes, and $N$ represents the number of elements per sample. ${p}_{ij}$ denotes the true label assigned by the model for the $i$-th sample belonging to $C$ class, while ${q}_{ij}$ signifies the model predicted probability for the $j$-th sample belonging to $C$ class.}

\section{Experiment}
Our method has been verified on the datasets of natural scene images CIFAR10-LT/100-LT\cite{krizhevsky2009learning}, STL10-LT\cite{coates2011analysis}, and the medical dataset TissueMNIST, and has been compared to the current prior arts. The experimental results show that UDTS achieves the performance improvements compared to other methods. Furthermore, UDTS can serve as a general method to be added to other methods and applied to other datasets.
%Furthermore, UDTS can serve as a versatile technique that can be integrated into various approaches and applied to diverse datasets.
\begin{figure*}[t]
    \centering
    \includegraphics[width=18cm]{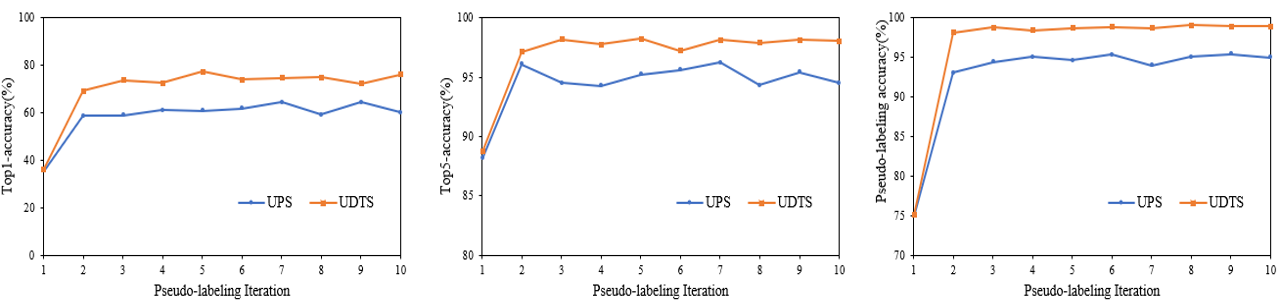}
    \caption{Comparison of UDTS and UPS\cite{rizve2021defense} in top-1, top-5, pseudo-labeling accuracy in different training iterations.}
    \label{UDTS vs UPS}
\end{figure*}
\subsection{Dataset}

We first conducted a comparison with a supervised baseline, training it on labeled data using the cross-entropy loss. Subsequently, we compared various semi-supervised methods, all built upon the foundation of FixMatch, which serves as a robust baseline. As the other methods for handling long-tailed data are also extensions of FixMatch, we implemented these extensions on top of FixMatch. To ensure the fairness of the experiment, we use the same backbone and the same super parameters to compare with the previous open-source methods: CREST\cite{wei2021crest}, CREST +\cite{wei2021crest}, DARP\cite{kim2020distribution}, DASO\cite{oh2022daso}, SAW\cite{lai2022smoothed}, DASH\cite{xu2021dash}, DEBIASPL\cite{wang2022debiased}. The results reported are based on the mean and standard deviation of three independent runs.

\begin{figure}[bp]
    \centering
    \includegraphics[width=8.5cm]{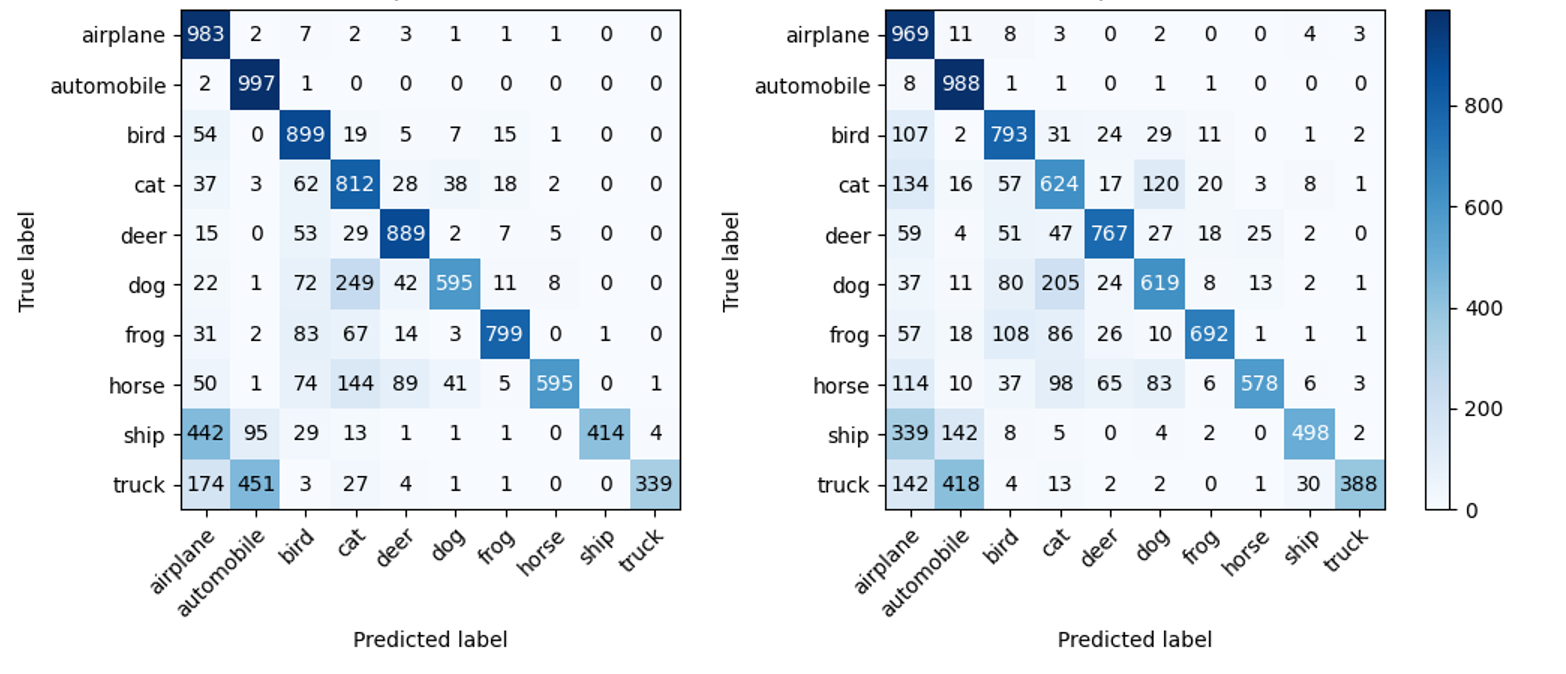}
    \caption{Comparison of FixMatch with UDTS in confusion matrix under the conditions of ${\gamma }_{lb}={\gamma }_{ulb}$=100, ${{D}_{lb}}$=1500 and ${{D}_{ulb}}$=3000 on CIFAR-10-LT.}
    \label{fig:confusion_matrix}
\end{figure}

\renewcommand\arraystretch{1.5}
\begin{table}
\normalsize   
\centering
\caption{Accuracy comparison with other methods on STL-10-LT.}
\vspace{0.75\baselineskip} 
\label{table2}
\setlength{\tabcolsep}{4pt}
%\resizebox{\textwidth}{45mm}
\begin{tabular}{lcc}\hline
\multirow{7}*{Algorithm} & \multicolumn{2}{c}{STL-10-LT} \\  
     &${{\gamma }_{lb}}$=10 ${{\gamma }_{ulb}}$=NA& ${{\gamma }_{lb}}$=20 ${{\gamma }_{ulb}}$=NA   \\ 
     \cmidrule(lr){2-2} \cmidrule(lr){3-3} 
     &$N$ = 150 & $N$ = 150 \\
     &$M$ = 100k & $M$ = 100k   \\
     \cmidrule(lr){1-1}\cmidrule(lr){2-3} 
Supervised   & 46.45${\pm}${\fontsize{8pt}{14pt}\selectfont 0.58}
  & 40.8${\pm}${\fontsize{8pt}{14pt}\selectfont0.64}  \\
FixMatch\cite{sohn2020fixmatch}    & 67.70${\pm}${\fontsize{8pt}{14pt}\selectfont2.02} & 56.90${\pm}${\fontsize{8pt}{14pt}\selectfont3.19}    \\
w/CReST\cite{wei2021crest}     & 66.28${\pm}${\fontsize{8pt}{14pt}\selectfont1.94} & 62.24${\pm}${\fontsize{8pt}{14pt}\selectfont2.16}   \\
w/CReST+\cite{wei2021crest}        & 66.40${\pm}${\fontsize{8pt}{14pt}\selectfont1.04} & 63.49${\pm}${\fontsize{8pt}{14pt}\selectfont1.86}  \\
w/DARP\cite{kim2020distribution}         & 64.56${\pm}${\fontsize{8pt}{14pt}\selectfont1.24} & 56.95${\pm}${\fontsize{8pt}{14pt}\selectfont2.76}    \\
% w/ ABC\cite{lee2021abc}         & 56.95${\pm}${\fontsize{8pt}{14pt}\selectfont1.2} & 67.68${\pm}${\fontsize{8pt}{14pt}\selectfont1.95}   \\
w/DASO\cite{oh2022daso}          & 71.13${\pm}${\fontsize{8pt}{14pt}\selectfont1.4} & 62.12 ${\pm}${\fontsize{8pt}{14pt}\selectfont4.05}   \\
w/SAW\cite{lai2022smoothed}         & 70.45${\pm}${\fontsize{8pt}{14pt}\selectfont0.71} & 66.42 ${\pm}${\fontsize{8pt}{14pt}\selectfont1.08}   \\
w/DASH\cite{xu2021dash}         & 70.58${\pm}${\fontsize{8pt}{14pt}\selectfont1.52} & 66.75 ${\pm}${\fontsize{8pt}{14pt}\selectfont0.89}    \\
w/Debiaspl\cite{wang2022debiased}         & 64.72${\pm}${\fontsize{8pt}{14pt}\selectfont0.98} & 56.23${\pm}${\fontsize{8pt}{14pt}\selectfont2.74}    \\
w/UDTS (ours)     & \textbf{71.24}${\pm}${\fontsize{8pt}{14pt}\selectfont0.58} 
& \textbf{66.86}${\pm}${\fontsize{8pt}{14pt}\selectfont0.37}    \\ \hline
\multicolumn{3}{c}{} 
\end{tabular}
\label{tab_stl}
\end{table}

CIFAR-10-LT/CIFAR-100-LT: CIFAR-10 and CIFAR-100\cite{krizhevsky2009learning} both contain 60,000 images, including 50,000 for training and 10,000 for testing, with 10 and 100 classes respectively. To ensure the accuracy and fairness of the experiment, we use CIFAR-10/100-LT under the same long-tailed setting. The CIFAR-10-LT dataset is conducted under the conditions of ${{D}_{lb}}=1500$, ${{D}_{ulb}}=3000$, ${{\gamma }_{lb}={\gamma }_{lb}=150}$, and ${{D}_{lb}}=500$, ${{D}_{ulb}}=4000$ and ${{\gamma }_{lb}={\gamma }_{lb}=100}$, respectively. The CIFAR-100-LT dataset is conducted under the conditions of ${{D}_{lb}}=150$, ${{D}_{ulb}}=1500$, ${{\gamma }_{lb}={\gamma }_{lb}=10}$, and ${{D}_{lb}}=150$, ${{D}_{ulb}}=300$, ${{\gamma }_{lb}={\gamma }_{lb}=15}$.

The STL-10 includes 113,000 RGB images with 96 $\times$ 96 resolutions, of which 5000 are in the training set, 8000 are in the test set, and the remaining 100,000 are unlabeled images. We take the conditions ${{\gamma }_{lb}=10}$, ${{D}_{lb}}=150$, ${{D}_{ulb}}=100k$ and ${{\gamma }_{lb}}=20$, ${{D}_{lb}}=150$, ${{D}_{ulb}}=100k$ to compare with other methods.

TissueMNIST\cite{yang2023medmnist} is a medical dataset of human kidney cortex cells, segmented from 3 reference tissue specimens and organized into 8 categories. The dataset consists of a total of 236,386 image samples are split with a ratio of 7 : 1 : 2 into training (165,466 images), validation (23,640 images) and test set (47,280 images). Each gray-scale image is 28 × 28 pixels. 

\renewcommand\arraystretch{1.5}
\begin{table}
\normalsize   
\centering
\caption{Accuracy comparison with other methods on TissueMNIST.}
\vspace{0.75\baselineskip} 
\label{table3}
\setlength{\tabcolsep}{4pt}
%\resizebox{\textwidth}{45mm}
\begin{tabular}{lcc}\hline
\multirow{7}*{Algorithm} & \multicolumn{2}{c}{TissueMNIST} \\  
     &${{\gamma }_{lb}}$=10 ${{\gamma }_{ulb}}$=NA& ${{\gamma }_{lb}}$=20 ${{\gamma }_{ulb}}$=NA   \\ 
     \cmidrule(lr){2-2} \cmidrule(lr){3-3} 
     &$N$ = 80 & $N$ = 400 \\
     &$M$ = 260k & $M$ = 200k   \\
     \cmidrule(lr){1-1}\cmidrule(lr){2-3} 
Supervised   & 40.09${\pm}${\fontsize{8pt}{14pt}\selectfont 2.93}
  & 45.90${\pm}${\fontsize{8pt}{14pt}\selectfont1.52}  \\
FixMatch\cite{sohn2020fixmatch}    & 44.05${\pm}${\fontsize{8pt}{14pt}\selectfont4.06} & 49.07${\pm}${\fontsize{8pt}{14pt}\selectfont1.23}    \\
MixMatch\cite{berthelot2019mixmatch}     & 44.27${\pm}${\fontsize{8pt}{14pt}\selectfont2.29} & \textbf{50.92}${\pm}${\fontsize{8pt}{14pt}\selectfont1.06}   \\
ReMixMatch\cite{berthelot2019remixmatch}        & 40.71${\pm}${\fontsize{8pt}{14pt}\selectfont5.16} & 47.08${\pm}${\fontsize{8pt}{14pt}\selectfont3.93}  \\
UDA\cite{kim2020distribution}         & 44.12${\pm}${\fontsize{8pt}{14pt}\selectfont3.26} & 43.05${\pm}${\fontsize{8pt}{14pt}\selectfont2.76}    \\
% w/ ABC\cite{lee2021abc}         & 56.95${\pm}${\fontsize{8pt}{14pt}\selectfont1.2} & 67.68${\pm}${\fontsize{8pt}{14pt}\selectfont1.95}   \\
FlexMatch\cite{zhang2021flexmatch}          & 42.77${\pm}${\fontsize{8pt}{14pt}\selectfont2.50} & 47.94 ${\pm}${\fontsize{8pt}{14pt}\selectfont1.78}   \\
CoMatch\cite{li2021comatch}         & 42.85${\pm}${\fontsize{8pt}{14pt}\selectfont3.46} & 48.17 ${\pm}${\fontsize{8pt}{14pt}\selectfont0.71}   \\
w/UDTS (ours)     & \textbf{45.33}${\pm}${\fontsize{8pt}{14pt}\selectfont3.05} 
& 48.07${\pm}${\fontsize{8pt}{14pt}\selectfont1.59}    \\ \hline
\multicolumn{3}{c}{} 
\end{tabular}
\label{tab_medmnist}
\end{table}

\renewcommand\arraystretch{1.5}
\begin{table}[t]
\centering
\caption{Experimental results on CIFAR10-LT with ${{\gamma }_{ulb}}=100$, ${{D}_{lb}}=1500$, and ${{D}_{ulb}}=3000$. UDTS gets higher top-1 and top-5 accuracy compared with UPS.}
\vspace{0.75\baselineskip} 
\label{tab_vs}
\setlength{\tabcolsep}{14pt}{
\scalebox{1.15}{
\begin{tabular}{l|cc}
\hline
Method     & Top1-acc ($\%$) & Top5-acc ($\%$) \\ \hline
Supervised & 46.63     & 83.26        \\
UPS\cite{rizve2021defense}        & 60.50     & 94.52     \\ \hline
UDTS       & \textbf{76.12}     & \textbf{98.02}     \\ \hline
\end{tabular}}}
\end{table}

\subsection{Implementation Details}

All experiments were conducted in PyTorch\cite{paszke2019pytorch}. Wide ResNet-28-2\cite{zagoruyko2016wide} is used as the network backbone. SGD optimizer is used to conduct model training based on $batch-size=64$, $momentum=0.99$, $weight\ decay\ factor = 0.0005$ and $learning\ rate = 0.03$. We use the same basic data expansion methods such as random resizing, random clipping, random horizontal flipping. In the uncertainty calculation, the Monte Carlo dropout drop-rate is set to 0.5 and makes $T=10$ predictions of the resulting uncertainty. 

\renewcommand\arraystretch{1.5}
\begin{table*}[t]
\centering
\caption{Accuracy results of ablation experiment on CIFAR10-LT, CIFAR100-LT, STL-10-LT, TissueMNIST. The experiments have three conditions: 1. when UDTS is selected without dynamic threshold. 2. when UDTS has no uncertainty-aware selection. 3. when uncertainty loss is introduced.}
\vspace{0.75\baselineskip}
\label{tab:abl_blending}
\setlength{\tabcolsep}{2pt}{
\scalebox{1.15}{
\begin{tabular}{lcccccc}
\hline
 \multirow{7}*{Algorithm}             &  \multicolumn{2}{c}{CIFAR10-LT}   &  \multicolumn{2}{c}{CIFAR100-LT}  &  \multicolumn{1}{c}{STL-10-LT}&  \multicolumn{1}{c}{TissueMNIST}     \\
            & ${{\gamma }_{lb}={\gamma }_{ulb}}$=100 & ${{\gamma }_{lb}={\gamma }_{ulb}}$=150 & ${{\gamma }_{lb}}$=${{\gamma }_{ulb}}$=10 & ${{\gamma }_{lb}}$=${{\gamma }_{ulb}}$=15 & ${{\gamma }_{lb}}$=20 ${{\gamma }_{ulb}}$=NA & ${{\gamma }_{lb}}$=10 ${{\gamma }_{ulb}}$=NA \\
            \cmidrule(lr){2-2}\cmidrule(lr){3-3}\cmidrule(lr){4-4}\cmidrule(lr){5-5}\cmidrule(lr){6-6}\cmidrule(lr){7-7}
    & $N$=1500 & $N$=500 & $N$=150 & $N$=150 & $N$=150 & $N$=80 \\ 
           & $M$=3000 & $M$=4000 & $M$=300 & $M$=300 & $M$=100k& $M$=260k       \\ 
            \cmidrule(lr){1-1}\cmidrule(lr){2-3} \cmidrule(lr){4-5}
            \cmidrule(lr){6-6}\cmidrule(lr){7-7}
FixMatch & 72.41     & 73.01    & 57.76 & 54.29 & 56.90 & 44.05   \\
UDTS, no selection & 74.14     &75.02  & 58.23 & 54.89 & 60.35 & \textbf{45.46}       \\
UDTS, no dynamic threshold & 76.51  &77.76  & 58.57 & 55.14 & 61.34 & 43.87      \\
UDTS, no uncertainty loss & 76.78  & 77.62 & 59.02  & 55.78 & 63.47 & 45.17   \\ \hline
UDTS, full method   & \textbf{78.13}    &\textbf{79.63}  & \textbf{59.82} & \textbf{56.28} & \textbf{66.86} & 45.33  \\ \hline
\label{ablation study}
\end{tabular}}}
\end{table*}

One of the main drawbacks of uncertainty estimation is that it necessitates multiple forward passes (denoted as $T$ times in the current work) to measure the uncertainty $u$. The computational overhead as well as total training time will significantly grow especially when dataset size is large or choosing a high $T$ value. Hence, we do an experiment on the number $T$ of forward propagation. We choose the value of $T$ as 10, taking into account the dataset size and the comprehensive consideration of multiple forward propagation on model performance and training time.

\renewcommand\arraystretch{1.5}
\begin{table}[t]
\centering
\caption{Accuracy experimental results on CIFAR10-LT with ${{\gamma }_{ulb}}=100$, ${{D}_{lb}}=1500$, and ${{D}_{ulb}}=3000$. The selection of hyperparameter T}
\vspace{0.75\baselineskip} 
\label{tab_T}
\setlength{\tabcolsep}{22pt}{
\scalebox{1.15}{
\begin{tabular}{lc}
\hline
Hyperparameter     & Top1-acc ($\%$)  \\ \hline
T=6 &  68.72             \\
T=8        &  72.46         \\ 
T=10      & \textbf{73.14}        \\ 
T=12      & 69.22          \\ \hline
\end{tabular}}}
\end{table}

\subsection{Comparisons to Prior Arts}

We compare UDTS with existing approaches, as well as the baseline network, on CIFAR-10-LT/CIFAR-100-LT\cite{krizhevsky2009learning}, STL-10-LT\cite{coates2011analysis}, TissueMNIST\cite{yang2023medmnist}, as shown in Table \ref{tab_cifar} and Table \ref{tab_stl}. Experimental results show that our method has good performance. Specifically, Figures \ref{fig: t-SNE} and \ref{fig:confusion_matrix} compare UDTS and FixMatch on CIFAR10-LT dataset by the recall, test accuracy, t-SNE, and confusion matrix. Figure \ref{fig: visualization} depicts the visualized experimental results of our approach compared to FixMatch.

UDTS screens incorrect pseudo-labels through uncertainty estimation, excluding pseudo-labels that negatively affect model performance, and ensures the accuracy of pseudo-labels through adaptive thresholding. Experimental results on three datasets also prove the effectiveness of UDTS. Although UDTS did not achieve SOTA results on medical images, experiments also confirm that its effectiveness and generalizability on medical datasets. It should be noted that we did not evaluate L2AC \cite{wang2022imbalanced}and InPL\cite{yu2023inpl} methods in our experiments, as their implementations differ from ours.

L2AC\cite{wang2022imbalanced} core idea is to automatically assimilate the training bias caused by class imbalance via the bias adaptive classifier, which is composed of a novel bias attractor and the original linear classifier. The bias attractor is designed as a light-weight residual
network and optimized through a bi-level learning framework. Such a learning strategy enables the bias adaptive classifier to fit imbalanced training data, while the linear classifier can provide unbiased label prediction for each class.
InPL\cite{yu2023inpl} takes the unlabeled sample to see if it was likely to be \textquotedblleft in-distribution\textquotedblright. To decide whether an unlabeled sample is \textquotedblleft in-distribution\textquotedblright or \textquotedblleft out-of-distribution\textquotedblright,
they adopt the energy score from out-of-distribution detection literature. Unfortunately, they don't have the open source code yet, so we can't compare this method in the same experimental setup. 
%Both of these methods are effective methods to solve the problem of semi-supervised long-tailed classification.

\noindent{{\bfseries UDTS vs UPS}}
We conducted experiments comparing UDTS to UPS \cite{rizve2021defense}, selecting pseudo-labels based on the uncertainty selection mechanism. These experiments were conducted on CIFAR10-LT dataset with the following settings:  ${{D}_{lb}=1500}$, ${{D}_{ulb}}=3000$, ${{\gamma }_{lb}=150}$, ${{\gamma }_{ulb}=150}$. The same backbone and parameter settings as UPS are used at the same time. Experimental results are shown in the Figure \ref{UDTS vs UPS} and Table \ref{tab_vs}. UDTS achieves better results than UPS with uneven data. Due to the influence of long-tailed data on model training, the uncertainty predicted for each class is different, and unlabeled data cannot be fully utilized through fixed threshold selection. We handle the problem of long-tailed data by selecting more reliable and diverse pseudo-labels through uncertain pseudo-label selection and adaptive uncertainty threshold. The experimental results show improvements over UPS in both top-1 and top-5 accuracy metrics. This demonstrates the superior applicability of UDTS to long-tailed data classification without the need for manual threshold adjustments based on the dataset.

\subsection{Ablation Study}

We conduct ablation studies to demonstrate the validity of the components of UDTS. Table \ref{ablation study} shows the results of the ablation experiments we conducted. It is evident that both the uncertainty-aware selection module and the uncertainty dynamic threshold algorithm have contributed to enhancing the network performance to some extent. The ablation studies demonstrate the effectiveness of each proposed modules for mitigating the challenges posed by class imbalance. 

When the experiment is set to CIFAR10-LT with ${{\gamma }_{lb}={\gamma }_{ulb}}=150$, ${{D}_{lb}}=1500$, and ${{D}_{ulb}}=3000$ and ${{\gamma }_{lb}={\gamma }_{ulb}}=100$, ${{D}_{lb}}=500$, and ${{D}_{ulb}}=4000$. When selecting by uncertainty using Monte Carlo dropout, the model accuracy improves by 4.62$\%$, but the improvement is constrained by the use of fixed thresholds for filtering. When the adaptive uncertainty threshold is added, it is observed that the model experiences an additional 1.87$\%$ improvement through the adaptive threshold selection mechanism. The introduction of uncertainty loss also improves the model performance by 2.61$\%$.

Due to the heterogeneity of pathological images and the stochasticity of network structures, ablation experiments on TissueMNIST may exhibit some randomness. This was demonstrated in the ablation study by removing the dynamic threshold, which was 0.18$\%$ lower than FixMatch. This is because, in the process of uncertainty-based selection, the heterogeneity of pathological images and their inherent characteristics may result in a higher level of uncertainty for most pathological images during model training. As a result, only a small number of pseudo-labels are selected, thereby impacting the training effectiveness.

\subsection{How to use UDTS?}
UDTS can function as a flexible technique that can be integrated into various methodologies and applied to a wide range of datasets. By incorporating Monte Carlo Dropout into a chosen backbone, such as ResNet50, and performing $T$ iterations for forward propagation, we can calculate the magnitude of uncertainty. Following this, in each iteration round, a process similar to the algorithm \ref{alg2} is introduced, which includes dynamic uncertainty threshold selection. Lastly, the fine-tuning hyperparameters based on the specific task and network variations allows for the application of UDTS.

\section{Conclusion}
In the current work, to alleviate the issue of model predictions being biased towards dominant classes caused by long-tailed data distribution in semi-supervised learning, we propose an Uncertainty-Aware Dynamic Threshold Selection (UDTS) approach which enables the model to dynamically adjust the selection thresholds for samples, thereby effectively mitigating the issue of long-tail data across training stages. For semi-supervised learning, UDTS facilitates dynamic and precise learning of long-tailed data characteristics, effectively preventing overfitting in predominant sample classes. The experimental results on the datasets of natural scene images CIFAR10-LT, CIFAR100-LT, STL-10-LT, and the dataset of medical images TissueMNIST empirically validate the effectiveness of UDTS.

\bibliographystyle{IEEEtran}
\bibliography{ref}

% \end{thebibliography}

\newpage

\vfill

\end{document}